\documentclass{bmvc2k}

\title{Multiple Fusion Adaptation:\\ A Strong Framework for Unsupervised Semantic Segmentation Adaptation}

\addauthor{Kai Zhang}{zhangkai193@mails.ucas.edu.cn}{1}
\addauthor{Yifan Sun}{sunyf15@tsinghua.org.cn}{2}
\addauthor{Rui Wang}{wangrui@iscas.ac.cn}{1}
\addauthor{Haichang Li}{haichang@iscas.ac.cn}{1}
\addauthor{Xiaohui Hu}{hxh@iscas.ac.cn}{1}

\addinstitution{
 Institute of Software,\\
 Chinese Academy of Sciences, China
}
\addinstitution{
 Baidu Research, China
}

\runninghead{ZHANG, Sun, Wang, Li, Hu}{Multiple Fusion Adaptation}

\usepackage{graphicx}
\usepackage{amsmath}
\usepackage{amsthm}
\usepackage{amsfonts}
\usepackage{wrapfig,lipsum,booktabs}

\usepackage{algorithm}
\usepackage{algorithmic}

\begin{document}

\maketitle

\begin{abstract}
This paper challenges the cross-domain semantic segmentation task, aiming to improve the segmentation accuracy on the unlabeled target domain without incurring additional annotation. Using the pseudo-label-based unsupervised domain adaptation (UDA) pipeline, we propose a novel and effective Multiple Fusion Adaptation (MFA) method. MFA basically considers three parallel information fusion strategies, \emph{i.e.}, the cross-model fusion, temporal fusion and a novel online-offline pseudo label fusion. Specifically, the online-offline pseudo label fusion encourages the adaptive training to pay additional attention to difficult regions that are easily ignored by offline pseudo labels, therefore retaining more informative details. While the other two fusion strategies may look standard, MFA pays significant efforts to raise the efficiency and effectiveness for integration, and succeeds in injecting all the three strategies into a unified framework. Experiments on two widely used benchmarks, \emph{i.e.}, GTA5-to-Cityscapes and SYNTHIA-to-Cityscapes, show that our method significantly improves the semantic segmentation adaptation, and sets up new state of the art (58.2\% and 62.5\% mIoU, respectively). The code will be available at~\url{https://github.com/KaiiZhang/MFA}.
\end{abstract}

\section{Introduction}

This paper considers the unsupervised domain adaptation (UDA) for semantic segmentation. In real-world segmentation tasks, there usually exists a domain gap between the training (source domain) and testing data (target domain), which substantially compromises the segmentation accuracy. Instead of using additional annotated data on the target domain for adaptation, which is notoriously expensive, an alternative way is to adapt the already-learned model through UDA ~\cite{hoffman2018cycada,RN180,RN162}. In another word, we aim to improve the segmentation accuracy on an unlabeled target domain without incurring additional annotation. 


A popular pipeline adopted by many state-of-the-art methods~\cite{RN180,RN132,zhang2021prototypical} consists of two training stages, \emph{i.e.}, a warm-up supervised training on the source domain and a sequential self-training on the target domain. Specifically, the first stage trains a warm-up model on the source domain data. For better generalization ability, the warm-up training process is typically assisted with some domain alignment constraints~\cite{RN166,RN135}. Then, the second training stage further adapts the warm-up model to the target domain through self-training~\cite{RN162, zou2019confidence}. The self-training usually uses the warm-up model to assign pseudo labels on the target domain, which are used to re-train (fine-tune) the model. 

\begin{figure}
\begin{center}
\includegraphics[width=0.6\textwidth]{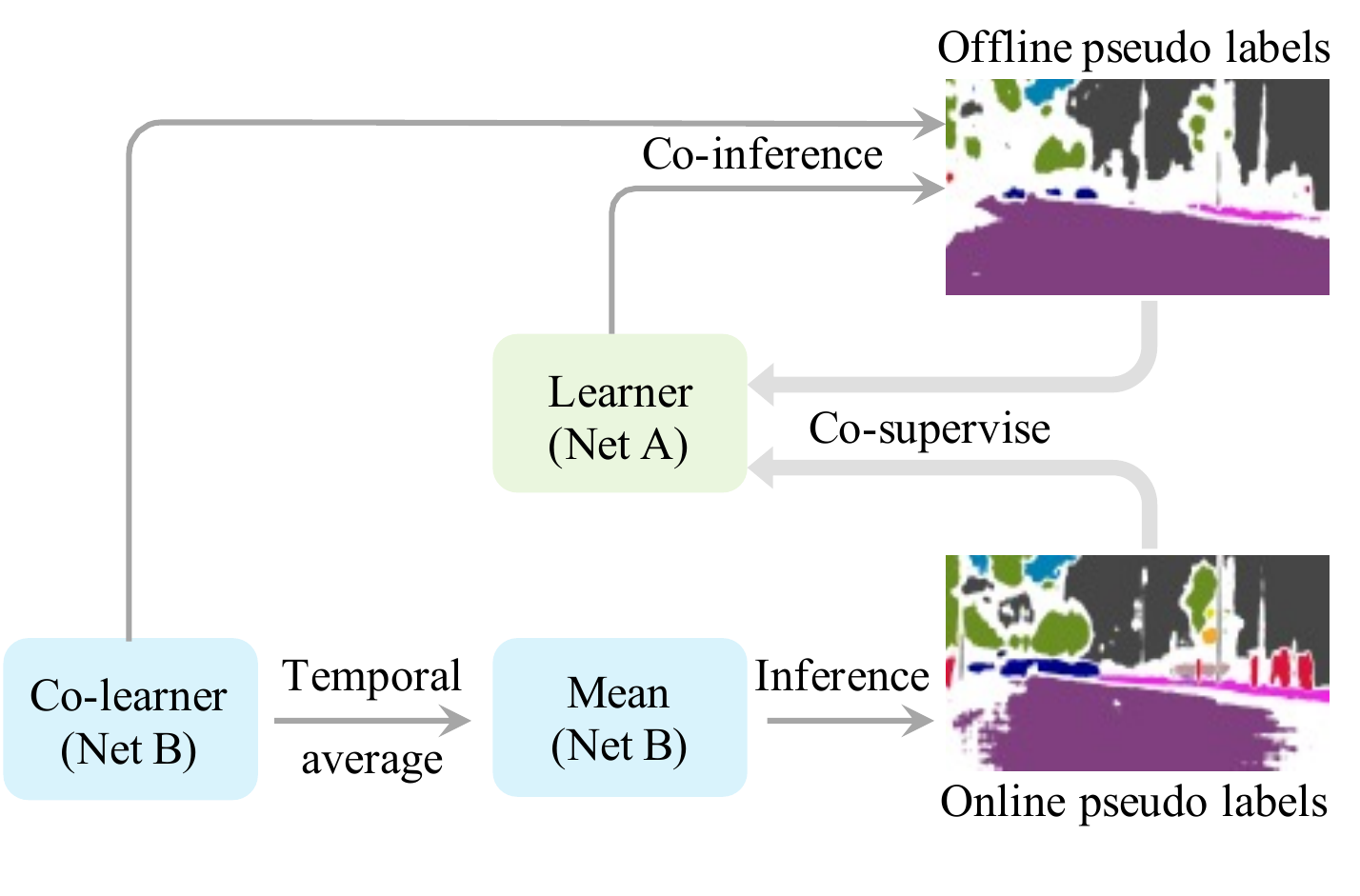}
\end{center}
\vspace{-20pt}
\caption{The proposed Multiple Fusion Adaptation (MFA) employs a co-learning framework to integrate three information fusions \emph{i.e.}, cross-model fusion, temporal fusion and online-offline pseudo label fusion. The learner (Net A) is co-supervised by the offline pseudo labels, as well as the online labels generated by it co-learner (Net B). To make the online label predictions more stable, MFA smooths the co-learner by temporal average (Mean Net B). Importantly, we design the online pseudo labels to be complementary to the offline pseudo labels, which promotes better fusion effect. In the co-learning framework, Net A and Net B will exchange their role of learner and co-learner. We only present Net A as the learner here for easier understanding. 
}
\label{fig: intro}
\end{figure}


This paper proposes a novel and effective Multiple Fusion Adaptation (MFA) method, based on the above-described two-stage UDA pipeline. We employ three basic information fusion to improve the domain adaptation, namely a novel \textbf{online-offline pseudo label fusion}, cross-model fusion and temporal fusion. In MFA, all the three fusion strategies are integrated in a co-learning framework, as illustrated in Figure~\ref{fig: intro}. Each learner is co-supervised by two types of pseudo labels, \emph{i.e.}, the online and the offline pseudo labels. The offline pseudo labels are generated with a popular method ~\cite{RN162}, while the online pseudo labels are generated by the temporal average model of the co-learner. This MFA pipeline has two advantages:

$\bullet$ \emph{The novel online-offline pseudo label fusion.} So far as we know, prior two-stage UDA methods~\cite{RN180,RN132} usually employ the offline pseudo labels. Among the iterations of ``assigning pseudo label'' and ``re-training'', the pseudo labels are updated after several training epochs, yielding the ``offline'' manner. While the offline manner allows additional post-processing and has the advantage of balanced pseudo labels~\cite{RN162}, it is prone to the ignorance of hard and informative samples \cite{RN204}. It is because the offline manner only preserves the most confident predictions among all the target domain data, which are relatively easy. As a remedy, we supplement the offline pseudo labels with online ones, which focus on the relatively hard details (\emph{i.e.}, the informative samples) within each training iteration. Moreover, since the online pseudo labels are generated by the up-to-date model (which has better accuracy than the historical ones), they reduce the exposure to noisy supervision and thus benefit the self-training process. 

$\bullet$ \emph{A highly efficient integration of three fusion strategies.} While the cross-model fusion and temporal fusion are quite popular, MFA pays significant efforts to raise the efficiency and effectiveness for integration, and succeeds in injecting all the three strategies into a unified framework. Specifically, MFA employs a co-learning framework consisted of two learners, as illustrated in Figure~\ref{fig: intro}. Each learner (model) in MFA is co-supervised by the offline pseudo labels generated by itself, as well as the online labels generated by its co-learner. To make the online predictions more stable, MFA smooths the co-learner by temporal average. Consequentially, MFA simultaneously enforces information fusion between 1) a model and its co-learner (and vice versa), 2) the up-to-date status and the temporal-averaged status and 3) online and offline pseudo labels. Combining these parallel fusions, MFA suppresses the pseudo label noises in the self-training stage and thus improves the segmentation adaptation.

Equipped with these two advantages, MFA is capable to improve UDA for semantic segmentation. First, the novel online-offline pseudo label fusion enables MFA to retain more informative details for adaptive training. Second, MFA manages a highly efficient integration of the online-offline pseudo label fusion and two commonly-adopted fusions, which further increases the accuracy of the pseudo labels. We evaluate the proposed MFA through extensive experiments. Experimental results show that MFA significantly improves the baseline and achieve performance on par with the state of the art. For example, on GTA5-to-Cityscapes and SYNTHIA-to-Cityscapes, MFA achieves $58.2\%$ and $62.5\%$ mIoU, respectively. Moreover, ablation study validates the effectiveness of each component in MFA. The main contributions of this paper are summarized as follows:
 
\begin{itemize}
\vspace{-2pt}
\setlength{\itemsep}{0pt}
\setlength{\parsep}{0pt}
\setlength{\parskip}{0pt}
    \item We propose MFA, an unsupervised semantic segmentation adaptation method based on self-training. MFA efficiently integrates three different information fusion strategies to improve the pseudo-label-based UDA. 
    \item Among the three fusion strategies of MFA, the online-offline pseudo label fusion is a novel one specifically designed for adaptive segmentation. The online pseudo labels supplement the self-training with relatively hard and informative samples, which may be easily ignored by the offline pseudo labels. 
    \item We conduct extensive experiments to evaluate the proposed MFA. Experimental results show that MFA achieves superior adaptive semantic segmentation and the training cost is relatively low. 
    
\end{itemize}

\section{Related work}

\paragraph{Semantic segmentation adaptation.} We divide the existing UDA semantic segmentation methods into two categories: domain alignment~\cite{hoffman2018cycada,RN180,RN138,RN135} and self-training~\cite{RN162,zou2019confidence,shin2020two,paul2020domain,pan2020unsupervised}, and the existing state-of-art approaches are usually a combination of two methods. The main motivation of domain alignment is to reduce the discrepancy between two domains. 
CyCADA~\cite{hoffman2018cycada} uses CycleGan~\cite{zhu2017unpaired} to transfer image style. FDA~\cite{RN180} proposes exchanging the low-frequency component of fourier transform without learning to achieve the same purpose. 
In addition, SIM~\cite{RN135} introduces the feature alignment of things and stuff respectively. 
In self-training, pseudo label learning~\cite{RN162} is a widely used approach. CBST~\cite{RN162} proposes an iterative pseudo label learning strategy and solve the class imbalance issue by class-independent confidence ranking. The improvement of pseudo label learning is an important research direction. In CRST~\cite{zou2019confidence}, a confidence regularized self-training method is proposed to address the problem of overconfident wrong label.~\cite{shin2020two} presents a two-phase pseudo label densification framework through voting-based and  easy-hard classification based method. In ~\cite{paul2020domain}, weak labels are explored to enhance pseudo label learning. Our work considers suppressing the pseudo label noise through multiple fusion strategies.

\paragraph{Temporal average.} Temporal ensembling~\cite{laine2016temporal} averages the outputs of the network-in-training to increase the prediction accuracy for the unlabelled samples. The mean teacher~\cite{tarvainen2017mean} averages model weights at different training steps to get a teacher model. The teacher model offers supervision signal through consistency constraint on the unlabeled samples. These works show that the temporal average of the deep model is more stable and accurate than the deep model at a single training step.

\paragraph{Learning with noisy labels.}  The information fusion between different models is an effective approach to suppress noises in labels. Co-teaching~\cite{han2018co} cross-trains two networks and let them teach each other given the possibly clean labels by small-loss trick. Co-teaching+~\cite{yu2019does} bridges the “Disagreement” strategy with the Co-teaching to enhance robustness under extremely noisy supervision. In these works, the labels are all available, which is different from the unsupervised domain adaptation problem. 

We note that~\cite{RN132,zhang2021prototypical} also consider the noise issue of pseudo labels on semantic segmentation adaptation. ~\cite{RN132} estimates the uncertainty of predictions and reduces the impact of low-confidence samples during pseudo label learning. \cite{zhang2021prototypical} take this issue by exploiting the feature distances from prototypes. Our method tackles the noisy pseudo label problem from a different viewpoint. We use co-learning and integrate multiple fusion strategies to resist the noisy pseudo labels, as well as to retain informative samples. Experimental results show that the proposed MFA marginally surpasses~\cite{RN132, zhang2021prototypical}.


\section{Approach}
\begin{figure}[t]
\centering
\includegraphics[width=1.0\linewidth]{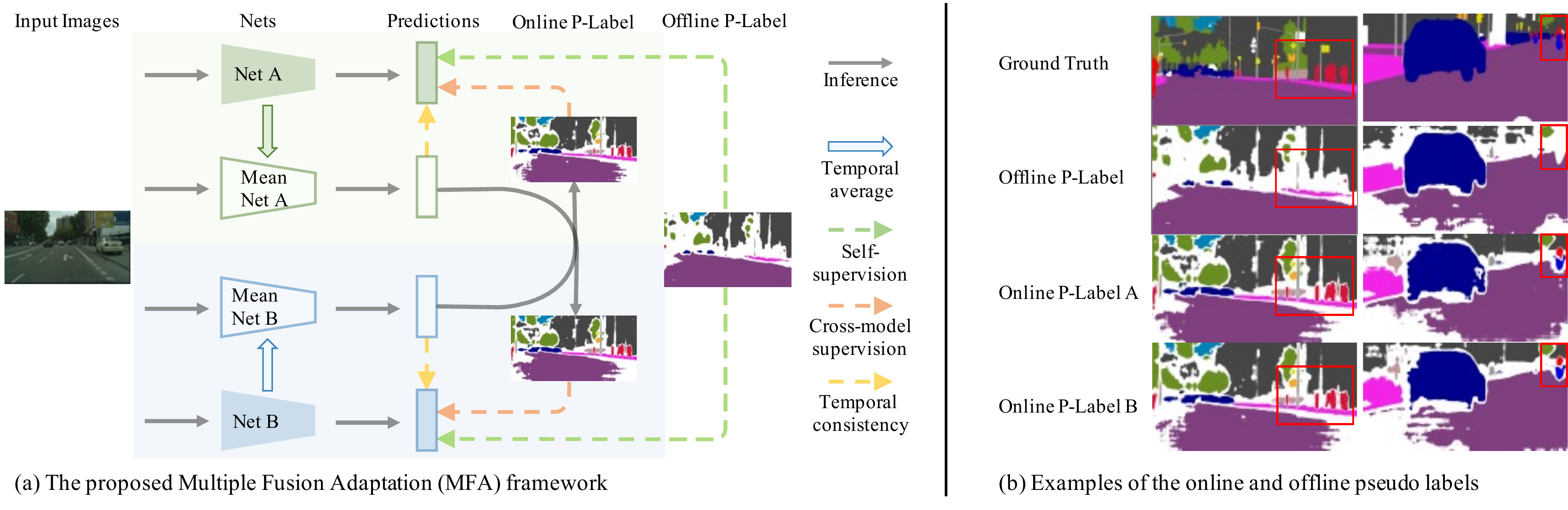}
\caption{\textbf{\textcolor{red}{(a)}} The overall framework of the proposed Multiple Fusion Adaptation (MFA). MFA collaboratively trains two models (Net A and Net B), with two supervision signals (Section \ref{sec: co-supervision}), \emph{i.e.}, the offline and online pseudo labels. The offline pseudo labels are generated by a popular baseline CBST \cite{RN162}, as introduced in Section \ref{sec: offline}. The online pseudo labels for Net A are generated by the temporal average of Net B (mean Net B), and vice versa (Section \ref{sec: online}).  
\textbf{\textcolor{red}{(b)}} Some examples of the online and offline pseudo labels. 
We highlight some regions with red bounding boxes to draw attention of the complementarity between these labels. For example, in the first column, the offline pseudo labels omit some important details about the pedestrians, and the online pseudo labels make up for this problem. In the second column, the person riding a motorcycle is recalled by online pseudo labels.}
\label{image:learn_step}
\end{figure}

The proposed MFA adopts the popular UDA pipeline of two-stage training, \emph{i.e.}, a warm-up training on the source domain and a following self-training on the target domain. We first give a formal description of the two-stage UDA pipeline in Section \ref{sec: pre}.
Based on this pipeline, MFA improves the adaptive segmentation through multiple fusions in the self-training stage, as illustrated in Figure~\ref{image:learn_step}. Basically, MFA uses two independent models (Net A and Net B) to set up a co-learning framework. Both models have the dual role of learner and co-learner. Before we collaboratively fine-tune them through self-training, we combine the two warm-up models to generate offline pseudo labels on the target domain (Section \ref{sec: offline}). During self-training, MFA smooths each model through temporal moving average and gets a corresponding ``mean net'' (\emph{i.e.}, Mean Net A and Mean Net B in Figure~\ref{image:learn_step}).  The function of Mean Net (or the temporal moving average operation) is two-fold. First, temporal moving average stabilizes the update of each mean net , therefore making the online predictions more stable. Second, according to the discovery in semi-supervised learning~\cite{tarvainen2017mean}, temporal moving average benefits from the ensemble of multiple models and thus maintains higher prediction accuracy. Given the current training mini-batch , each mean net predicts a respective set of online pseudo labels (Section \ref{sec: online}). Finally, MFA enforces co-supervision on each model. Specifically, each learner is co-supervised by the offline pseudo labels, as well as the online ones generated by its co-learner (Section \ref{sec: co-supervision}).
 
\subsection{Preliminaries on Two-stage UDA} \label{sec: pre}
In domain adaptive segmentation, we have two datasets belonging to different domains, \emph{i.e.}, the source domain and the unlabeled target domain. The source domain dataset is denoted as $D_{S}=\left \{ x_{S}^{i}, y_{S}^{i} \right \}_{i=1}^{N_{S}}$, where $x_{S}\in\mathbb{R}^{H \times W \times 3}$ is a color image in source domain, $N_{S}$ is the number of source data and $y_{S} \in \mathbb{R}^{H \times W}$ is the corresponding semantic map. The definition of the target domain dataset $D_{T}=\left \{ x_{T}^{j}, y_{T}^{j} \right \}_{j=1}^{N_{T}}$ is similar, except that $y_{T}$ is unknown. Let $F$ represents a semantic segmentation network, and $\theta$ stands the parameters of $F$. The goal of the UDA problem is to estimate $\theta$ to minimize the prediction error on the unlabeled target domain.


In two-stage UDA, the parameter $\theta$ of the warm up model are first obtained through training on the source domain (\emph{i.e.}, stage 1). Then in the self-training (\emph{i.e.}, stage 2), given the input sample from target domain, the prediction $\hat{y}_{T} = F\left ( x_{T} \mid \theta\right )$ is the predicted class probability map, where $\hat{y}_{T}\in\mathbb{R}^{H \times W \times C}$  and $C$ is the number of classes. And $\mathop{\max}\left(\hat{y}_{T}^{c}\right)\in\mathbb{R}^{H \times W}$ is the prediction confidence map. The one-hot map of pseudo labels is obtained by:%
\begin{equation}
\label{equ:option5}
\hat{P}\left ( x_{T} \mid \theta \right )=\text{one-hot}\left(\mathop{\arg\max}_{c}F\left ( c \mid x_{T}, \theta\right)\right)
\end{equation}%
In the standard self-training strategy, an optional way is to retrain the initialized model multiple times by merging the pseudo label data with the source data, which is applied in~\cite{RN135,RN180}. However, this strategy needs to reinitialize $F$ to start training, which is very time-consuming. Therefore, we choose another way used in~\cite{RN132} as our baseline, \emph{i.e.}, fine-tune the warm-up model on the pseudo labels. The loss function for self-supervision is formulated as follows:%
\begin{equation}
\label{equ:option2}
\mathcal{L}_{self}(x_{T},\theta) =-\mathop{\sum}_{batch}m \cdot \hat{p}_{T} \cdot \log\left(F\left ( x_{T} \mid\theta\right )\right)
\end{equation}%
Where $\hat{p}_{T}$ is obtained by $x_{T}$ and  $\theta$ in Equation~\ref{equ:option5}. And $m \in \mathbb{R}^{H \times W}$ is a binary mask for filtering out the unreliable pseudo labels. Specifically, if $m_{h, w}=1$, then we have $\hat{P}_{T, h, w}$ selected for training. In contrast, if $m_{h, w}=0$, the corresponding pseudo label is regarded as unreliable and thus ignored.

\subsection{Offline Pseudo Label}\label{sec: offline}
In Equation~\ref{equ:option2}, the pseudo labels are typically selected in an offline manner, \emph{i.e.}, they will not be instantly updated during model optimization. 
MFA combines two different warm-up models for offline pseudo label prediction $\tilde{y}_{T}$. 
Consequentially, the offline pseudo labels benefit from model ensemble. Given the raw predictions, we use the CBST~\cite{RN162} method to select the training samples (pixels) by:%
\begin{equation}
\label{equ:option7}
m = 
\begin{cases}
1  & \text{ if }\, c=\mathop{\arg\max}\left ( \tilde{y}_{T}^{c} \right )\, \& \, \mathop{\max}\left(\tilde{y}_{T}^{c}\right)>\tau_{c} \\
0 & otherwise
\end{cases}
\end{equation}%
where $c\in \left [ 0, C\right )$ is the class index, and $\tau_{c}$ is the thresholds for the corresponding class. Following~\cite{RN162}, we set each threshold $\tau_{c}$ to ensure class balance, \emph{i.e.}, all the classes has an identical proportion of the selected pixels.

The offline manner of CBST has the advantage of stabilizing supervision signals and avoiding the gradual dominance of large classes~\cite{RN162}. In MFA, the ensemble of two warm-up models further benefits the accuracy of the pseudo labels. However, the offline manner is prone to the problems of \emph{ignorance of hard samples} \cite{RN204} and \emph{longer exposure to the potential noisy labels}. We thus introduce the online pseudo labels as a supplementary.

\subsection{Online Pseudo Label} \label{sec: online}
\paragraph{Temporal average.} MFA instantly assigns online pseudo labels to images in the current mini-batch. In another word, the online pseudo labels do NOT require a post-processing over the whole target domain samples. To alleviate the potential instability issue, MFA first smooths each model through temporal average to generate mean net, which is formulated as:
\begin{equation}
\label{equ:option3}
\theta_{t}^{mean} = \alpha \theta_{t-1}^{mean}+(1-\alpha)\theta_{t}
\end{equation}%
Where $\alpha$ is a smoothing coefficient hyper-parameter, $\theta$ denotes the model parameters and $t$ is the training step. Given the mean net with parameters $\theta^{mean}$ and an input image $x_T$, MFA generates the corresponding online pseudo labels by Equation~\ref{equ:option5}.

\begin{wrapfigure}{R}{0.55\textwidth}
 \vspace{-9pt}
    \begin{minipage}{0.55\textwidth}
      \begin{algorithm}[H]
        \caption{Online CBST}
        \label{alg:cbst}
        \textbf{Input}: Mean Net $F(\theta_{A})$ and $F(\theta_{B})$, minibatch target data $x_{b}$. \\
        \textbf{Parameter}: Ratio $ \varphi $ of selected pseudo labels.\\
        \textbf{Output}: $LP_{b}^{1}$ and $LP_{b}^{2}$ from $\theta_{A}$ and $\theta_{B}$, respectively.
        \begin{algorithmic}
            \STATE $P_{b}^{1} = F\left ( x_{b} \mid \theta_{A} \right )$
            \STATE $P_{b}^{2} = F\left ( x_{b} \mid \theta_{B} \right )$
            \STATE $LP_{b}^{1} = \arg\max\left ( P_{b}^{1},axis=1\right ) $
            \STATE $LP_{b}^{2} = \arg\max\left ( P_{b}^{2},axis=1\right ) $
            \FOR {i=1\ to\ 2}
            \STATE $MP_{b}^{i} = \max\left ( P_{b}^{i},axis=1\right ) $
            \FOR {c=1\ to\ C}
            \STATE $MP_{c, b}^{i} = MP_{b}^{i}\left [ LP_{b}^{i}==c \right ]$
            \STATE $M_{c}^{i} = sort\left ( MP_{c, b}^{i}, order=descending \right ) $
            \STATE $len_{c}^{i} = length(M_{c}^{i})\times \varphi$
            \STATE $\tau_{c}^{i} = M_{c}^{i}\left [ len_{c}^{i} \right ] $
            \STATE $LP_{b}^{i}\left[ LP_{b}^{i}==c \ \& \ LP_{b}^{i} < \tau_{c}^{i}  \right] = 255$
            \ENDFOR
            \ENDFOR
            \STATE \textbf{return} $LP_{b}^{1}$, $LP_{b}^{2}$
        \end{algorithmic}
      \end{algorithm}
    \end{minipage}
\end{wrapfigure}

\paragraph{Online CBST.} The online pseudo labels require to be filtered as well, so as to remove the labels with relatively low confidence. To this end, we propose a novel Online-CBST. 

It is similar to the original CBST, except that it uses the data in current mini-batch (rather than the whole training data) as the reference for label selection. In other words, we set a respective filtering threshold for each class, so that all the classes have equal proportion of pseudo-labeled samples in current mini-batch. In analogy to the original CBST, we design the Online-CBST to have linearly-increasing proportion $\varphi(t) \in \left[\rho_{min}, \rho_{max} \right]$, which is the desired proportion in the $t-$th training step. $\rho_{min}$ and $\rho_{max} $ are the pre-defined minimum and maximum proportions. 
Initially, the accuracy of each warm-up model is relatively low. So we use a small $\varphi(0)=\rho_{min}$ to retain the most confident pseudo labels and abandon the others.

Since the online pseudo labels are generated by the up-to-date model, they reduce the exposure to noisy supervision. Moreover, Online CBST is beneficial for recalling the relatively hard details (\emph{i.e.}, the informative samples) within each training iteration. In Section~\ref{sec:Role of Online}, we analyzed this in more detail. Given the online pseudo labels $P(x_T\mid\theta_A^{mean})$ and $P(x_T\mid\theta_B^{mean})$, the Online-CBST correspondingly generates two masks $m_A$ and $m_B$ for selecting the pixels. The pseudo code for generating the online labels is to be accessed in Algorithm~\ref{alg:cbst}. 

\subsection{Co-supervision in MFA} \label{sec: co-supervision}

Given both the offline and online pseudo labels, MFA enforces a co-supervision on each learner in Figure~\ref{image:learn_step} (a). We illustrate the loss functions for such co-supervision as follows.

To cooperate with the offline pseudo labels, MFA uses the loss function $\mathcal{L}_{self}$ defined by Equation~\ref{equ:option2} for self-supervision on both Net A and Net B. As for the online pseudo labels, MFA uses $P(x_T\mid\theta_A^{mean})$ (predictions from Net A) for supervising Net B and uses $P(x_T\mid\theta_B^{mean})$ for supervising Net A, yielding the so-called cross-model supervision. The detailed loss functions are formulated as:%
\begin{equation}
\label{equ:option9}
\begin{aligned}
&\mathcal{L}_{cross}(\theta_{A}, \theta_{B}^{mean}) = -\mathop{\sum}_{batch}m_{\text{B}}\cdot P\left ( x_{T} \mid \theta_{B}^{mean} \right )   \cdot \log \left (F\left ( x_{T} \mid\theta_{A}\right )\right ) \\
& \mathcal{L}_{cross}(\theta_{B}, \theta_{A}^{mean}) = -\mathop{\sum}_{batch}m_{\text{A}}\cdot P\left ( x_{T} \mid \theta_{A}^{mean} \right ) \cdot \log \left (F\left ( x_{T} \mid\theta_{B}\right ) \right ) 
\end{aligned}
\end{equation}%
in which $m_B$ ($m_A$) is the selection-mask for $P(x_T\mid\theta_B^{mean})$ ($P(x_T\mid\theta_A^{mean})$) generated by the proposed Online-CBST.

Besides the co-supervision with online and offline pseudo labels, MFA enforces a respective consistency constraint between each learner and its temporal average. Similar to mean teacher~\cite{tarvainen2017mean}, the consistency loss is the expected distance between the prediction of the model and the prediction of the temporal average model, which is formulated as:%
\begin{equation}
\label{equ:option4}
\mathcal{L}_{cst}(\theta, \theta^{mean}) = \mathbb{E}_{x}\left [ \left \| F\left ( x_{T} \mid\theta^{mean}\right ) - F\left ( x_{T} \mid \theta\right ) \right \| \right ]
\end{equation}%
In summary, MFA sums up all the losses to collaboratively train Net A and Net B by:%
\begin{equation}
\label{equ:option10}
\begin{aligned}
\mathcal{L}_{all} = &\mathcal{L}_{self}\left(x_{T},\theta_{A}\right)+\mathcal{L}_{self}(x_{T},\theta_{B}) \\
 &+ \lambda_{cst} \left(\mathcal{L}_{cst}\left(\theta_{A},\theta_{A}^{mean}\right)+\mathcal{L}_{cst}\left(\theta_{B},\theta_{B}^{mean}\right)\right) \\
 &+  \lambda_{cross}\left(\mathcal{L}_{cross}\left(\theta_{A}, \theta_{B}^{mean}\right)+\mathcal{L}_{cross}\left(\theta_{B}, \theta_{A}^{mean}\right)\right),
\end{aligned}
\end{equation}%
in which $\lambda_{cst}$  and $\lambda_{cross}$  are the weighting factors for $\mathcal{L}_{cst}$  and $\mathcal{L}_{cross}$, respectively.

\section{Experiments}
\subsection{Datasets}
We evaluate the proposed MFA under two widely adopted cross-domain segmentation settings, \emph{i.e.}, GTA5-to-Cityscapes and SYNTHIA-to-Cityscapes. 
GTA5 and SYNTHIA are both synthetic datasets. The GTA5~\cite{richter2016playing} dataset consists of 24,966 synthesized images of resolution $1914 \times 1052$. Same as existing works, we evaluate our method on 19 common categories shared by GTA5 and Cityscapes. The SYNTHIA~\cite{ros2016synthia} dataset has 9,400 synthesized images of resolution $1280 \times 720$ with fine annotations. Following~\cite{RN135,RN180}, we report the per-class IoU and mIoU on the 13 common categories shared by SYNTHIA and Cityscapes. 

Cityscapes is a real-world semantic segmentation dataset~\cite{cordts2016cityscapes}, which consists of $5,000$ images of resolution $2048 \times 1024$ with pixel-level annotations. 
It is split into a training set, validation set and test set with $2,975$, $500$ and $1,525$ images, respectively. In line with the standard evaluation setting, we use the $2,975$ training images (without the ground-truth labels) as target domain images, and then evaluate the domain adaptive segmentation accuracy on the validation set. 

\subsection{Implementation Details} \label{section:architecture}
Following \cite{RN135,RN180}, we use DeepLabV2~\cite{chen2017deeplab} based on ResNet101 \cite{he2016deep} as the backbone model. We recall that MFA is based on the two-stage UDA pipeline and requires warm-up training. To promote the divergence between the initial status of two learners, we adopt two state-of-the-art domain alignment methods proposed by FDA~\cite{RN180} and SIM~\cite{RN135}. 
These two methods serves as the strong baseline for MFA. That being said, we will show that MFA achieves significant improvement over these (\emph{e.g.}, +10.1\% mIoU on GTA5-to-Cityscapes), which results in the state-of-the-art performance. Moreover, MFA is compatible to any warm-up models and is potential to benefit from the future progress in domain alignment. 

We use SGD optimization strategy with momentum 0.9. We initialize the learning rate to $2e^{-4}$, and adjust it according to the poly learning rate scheduler with a power of 0.9. As for the hyper-parameters, Equation~\ref{equ:option3} has $\alpha = 0.99$ for temporal average, and Online-CBST has $\rho_{min}=0.2$, $\rho_{max}=0.7$. Moreover, we set $\lambda_{cst}=1.0$, $\lambda_{cross}=0.5$ for Equation~\ref{equ:option10}.

\begin{table}
\begin{center}
\resizebox{\textwidth}{20mm}{
\begin{tabular}{|l|cccccccccccccccccccc|}
\hline
Method & \rotatebox{90}{road} & \rotatebox{90}{sdwk} & \rotatebox{90}{bldng} & \rotatebox{90}{wall} & \rotatebox{90}{fence} & \rotatebox{90}{pole} & \rotatebox{90}{light} & \rotatebox{90}{sign} & \rotatebox{90}{veg} & \rotatebox{90}{trrn} & \rotatebox{90}{sky} & \rotatebox{90}{psn} & \rotatebox{90}{rider} & \rotatebox{90}{car} & \rotatebox{90}{truck} & \rotatebox{90}{bus} & \rotatebox{90}{train} & \rotatebox{90}{moto} & \rotatebox{90}{bike} & \rotatebox{90}{mIoU} \\
\hline\hline
AdaStruct~\cite{RN133} & 86.5 & 25.9 & 79.8 & 22.1 & 20.0 & 23.6 & 33.1 & 21.8 & 81.8 & 25.9 & 75.9 & 57.3 & 26.2 & 76.3 & 29.8 & 32.1 & 7.2 & 29.5 & 32.5 & 41.4     \\
Cycada~\cite{hoffman2018cycada} & 86.7 & 35.6 & 80.1 & 19.8 & 17.5 & 38.0 & 39.9 & 41.5 & 82.7 & 27.9 & 73.6 & 64.9 & 19.0 & 65.0 & 12.0 & 28.6 & 4.5 & 31.1 & 42.0 & 42.7 \\
WSDA~\cite{paul2020domain} & 91.6 & 47.4 & 84.0 & 30.4 & 28.3 & 31.4 & 37.4 & 35.4 & 83.9 & 38.3 & 83.9 & 61.2 & 28.2 & 83.7 & 28.8 & 41.3 & 8.8 & 24.7 & 46.4 & 48.2 \\
BDA~\cite{RN166} & 91.0 & 44.7 & 84.2 & 34.6 & 27.6 & 30.2 & 36.0 & 36.0 & 85.0 & 43.6 & 83.0 & 58.6 & 31.6 & 83.3 & 35.3 & 49.7 & 3.3 & 28.8 & 35.6 & 48.5 \\
SIM~\cite{RN135} & 90.6 & 44.7 & 84.8 & 34.3 & 28.7 & 31.6 & 35.0 & 37.6 & 84.7 & 43.3 & 85.3 & 57.0 & 31.5 & 83.8 & \textbf{42.6} & 48.5 & 1.9 & 30.4 & 39.0 & 49.2     \\
Seg-U~\cite{RN132} & 90.4 & 31.2 & 85.1 & 36.9 & 25.6 & 37.5 & \textbf{48.8} & \textbf{48.5} & 85.3 & 34.8 & 81.1 & 64.4 & 36.8 & 86.3 & 34.9 & 52.2 & 1.7 & 29.0 & 44.6 & 50.3\\
FDA~\cite{RN180} & 92.5  & 53.3  & 82.4  & 26.5  & 27.6  & 36.4  & 40.6  & 38.9  & 82.3  & 39.8  & 78.0  & 62.6  & 34.4  & 84.9  & 34.1  & 53.1  & 16.9  & 27.7  & 46.4  & 50.5\\
TPLD~\cite{shin2020two} & 94.2 & 60.5 & 82.8 & 36.6 & 16.6 & 39.3 & 29.0 & 25.5 & 85.6 & \textbf{44.9} & 84.4 & 60.6 & 27.4 & 84.1 & 37.0 & 47.0 & \textbf{31.2} & 36.1 & 50.3 & 51.2 \\
ProDA~\cite{zhang2021prototypical} & 91.5 & 52.4 & 82.9 & \textbf{42.0} & \textbf{35.7} & 40.0 & 44.4 & 43.3 & \textbf{87.0} & 43.8 & 79.5 & 66.5 & 31.4 & 86.7 & 41.1 & 52.5 & 0.0 & 45.4 & 53.8 & 53.7 \\
MFA(ours)  & \textbf{94.5} & \textbf{61.1} & \textbf{87.6} & 41.4 & 35.4 & \textbf{41.2} & 47.1 & 45.7 & 86.6 & 36.6 & \textbf{87.0} & \textbf{70.1} & \textbf{38.3} & \textbf{87.2} & 39.5 & \textbf{54.7} & 0.3 & \textbf{45.4} & \textbf{57.7} & \textbf{55.7}     \\
\hline
ProDA*~\cite{zhang2021prototypical} & 87.8 & 56.0 & 79.7 & 46.3 & 44.8 & 45.6 & 53.5 & 53.5 & 88.6 & 45.2 & 82.1 & 70.7 & 39.2 & 88.8 & 45.5 & 59.4 & 1.0 & 48.9 & 56.4 & 57.5 \\
MFA*(ours) & 93.5 & 61.6 & 87.0 & 49.1 & 41.3 & 46.1 & 53.5 & 53.9 & 88.2 & 42.1 & 85.8 & 71.5 & 37.9 & 88.8 & 40.1 & 54.7 & 0.0 & 48.2 & 62.8 & \textbf{58.2} \\

\hline
\end{tabular}}
\end{center}
\caption{ Results on GTA5-to-Cityscapes. MFA surpasses all the competing methods. For a fair comparison, ``*'' indicates additional distillation stage is used, which is proposed in~\cite{zhang2021prototypical}. }
\label{tab:result_gta5}
\vspace{-0.2cm}
\end{table}

\subsection{The Effectiveness of MFA}
Table~\ref{tab:result_gta5} compares MFA against several state-of-the-art UDA methods on the GTA5-to-Cityscapes benchmark, from which we draw two observations.
First, comparing MFA against all the competing two stage methods, we find that MFA surpasses all the competing methods by a large margin. 
For example, it achieves $55.7\%$ mIoU, which is higher than the strongest competitor ProDA by $+2.0\%$. We achieve the best scores on most categories (11 in 19). Second, under the fair comparison, MFA* presents $58.2\%$ mIoU, with additional model distillation stage~\cite{zhang2021prototypical}. In line with ~\cite{zhang2021prototypical}, we use the SimCLRv2~\cite{chen2020big} pretrained weights and same distillation strategy.


We also compare MFA with competing methods on the SYNTHIA-to-Cityscapes benchmark in the Table~\ref{tab:result_syn} and draw consistent observations as on GTA5-to-Cityscapes. MFA achieves higher mIoU than prior state-of-the-art methods. We report $58.7\%$ mIoU (after self-training stage) and $62.5\%$ mIoU (after distillation stage) on this benchmark. 

\subsection{The Efficiency of MFA}
The basic self-training strategy needs iteration of ``assigning pseudo label'' and ``re-training'', which is adopted by~\cite{RN162,RN180}. Benefit from the online-offline fusion, MFA converges faster without multiple iterations and reinitialization. The training in MFA lasts 65 epochs, and is more efficient than other methods~\cite{RN132,RN180,zhang2021prototypical,RN162}. We note that the warm-up models in FDA~\cite{RN180} and the proposed MFA achieve close performance, but MFA achieves 55.7 mIoU after self-training stage (compared to 50.5 in FDA after two round self-training stage). We thus infer that the superiority of MFA is due to the well-engineered self-training with multiple information fusion.

\subsection{The Benefit of Online Pseudo Labels} \label{sec:Role of Online}
We visualize some examples of online and offline pseudo labels in Figure~\ref{image:learn_step} (b). It is clearly observed that online pseudo labels are complementary to the offline ones and focus on the relatively hard details. In the first column, the offline pseudo labels omit the pedestrians, which look small in the image. In contrast, the online pseudo labels succeed in pointing the existence of these pedestrians. In the second column, the person riding a motorcycle is omitted by the offline pseudo labels and is recalled by online pseudo labels. We owe this to two reasons. First, the mean net is updated in time, which is conducive to producing more high-quality pseudo labels along with self-training. Second, since the proposed online CBST focuses on current batch of samples, the relatively difficult samples are more likely to be recalled in the confidence ranking.

\begin{table}
\begin{center}
\resizebox{\textwidth}{20mm}{
\begin{tabular}{|l|cccccccccccccc|}
\hline
Method & \rotatebox{90}{road} & \rotatebox{90}{sdwk} & \rotatebox{90}{bldng} & \rotatebox{90}{light} & \rotatebox{90}{sign} & \rotatebox{90}{veg} & \rotatebox{90}{sky} & \rotatebox{90}{psn} & \rotatebox{90}{rider} & \rotatebox{90}{car} & \rotatebox{90}{bus} & \rotatebox{90}{moto} & \rotatebox{90}{bike} & \rotatebox{90}{mIoU} \\
\hline\hline
AdaStruct~\cite{RN133} & 84.3 & 42.7 & 77.5 & 4.7 & 7.0 & 77.9 & 82.5 & 54.3 & 21.0 & 72.3 & 32.2 & 18.9 & 32.3 & 46.7 \\
BDA~\cite{RN166} & 86.0 & 46.7 & 80.3 & 14.1 & 11.6 & 79.2 & 81.3 & 54.1 & 27.9 & 73.7 & 42.2 & 25.7 & 45.3 & 51.4 \\
WSDA~\cite{paul2020domain} & \textbf{92.0} & \textbf{53.5} & 80.9 & 3.8 & 6.0 & 81.6 & \textbf{84.4} & 60.8 & 24.4 & 80.5 & 39.0 & 26.0 & 41.7 & 51.9\\
SIM~\cite{RN135}  & 83.0 & 44.0 & 80.3 & 17.1 & 28.7 & 15.8 & 81.8 & 59.9 & 33.1 & 70.2 & 37.3 & 28.5 & 45.8 & 52.1\\
TPLD~\cite{shin2020two} & 80.9 & 44.3 & 82.2 & 20.5 & 30.1 & 77.2 & 80.9 & 60.6 & 25.5 & 84.8 & 41.1 & 24.7 & 43.7 & 53.5 \\
Seg-U~\cite{RN132} & 87.6 & 41.9 & 83.1 & 31.3 & 19.9 & 81.6 & 80.6 & 63.0 & 21.8 & 86.2 & 40.7 & 23.6 & 53.1 & 54.9 \\
FDA~\cite{RN180}  & 79.3 & 35.0 & 73.2 & 19.9 & 24.0 & 61.7 & 82.6 & 61.4 & 31.1 & 83.9 & 40.8 & \textbf{38.4} & 51.1 & 52.5     \\
ProDA~\cite{zhang2021prototypical} &87.1 & 44.0 & 83.2 & \textbf{45.8} & \textbf{34.2} & \textbf{86.7} & 81.3 & 68.4 & 22.1 & \textbf{87.7} & \textbf{50.0} & 31.4 & 38.6 & 58.5 \\
MFA(ours) & 85.4 & 41.9 & \textbf{84.1} & 22.2 & 23.9 & 83.6 & 80.7 & \textbf{71.5} & \textbf{35.8} & 86.6 & 47.6 & 37.2 & \textbf{62.5} & \textbf{58.7}\\
\hline
ProDA*~\cite{zhang2021prototypical} & 87.8 & 45.7 & 84.6 & 54.6 & 37.0 & 88.1 & 84.4 &  74.2 &  24.3 & 88.2 & 51.1 & 40.5 & 45.6 & 62.0 \\
MFA*(ours)  & 81.8 & 40.2 & 85.3 & 38.0 & 33.9 & 82.3 & 82.0 & 73.7 & 41.1 & 87.8 & 56.6 & 46.3 & 63.8 & \textbf{62.5} \\
\hline
\end{tabular}}
\end{center}
\caption{Results on SYNTHIA-to-Cityscapes. MFA achieves better performance than the other state-of-the-art methods. The symbol ``*'' indicates additional distillation stage is used. }
\label{tab:result_syn}
\vspace{-0.2cm}
\end{table}

\subsection{Ablation Study}

\begin{wraptable}{r}{0.5\linewidth}
\centering
\vspace{-10pt}
\scalebox{0.8}{
\begin{tabular}{|l|ccc|cc|}
\hline
 Method & TF & CMF & OOF & mIoU & Gain \\
\hline
 Warm & & & &  45.6  & \\
 ST & & & & 49.4 & 3.8 \\
\hline
TF & \checkmark & & &  50.6 & 5.0 \\
TF$\&$CMF & \checkmark & \checkmark & & 51.7 & 6.1 \\
TF$\&$OOF  & \checkmark & & \checkmark &  52.7 & 7.1 \\
MFA & \checkmark & \checkmark & \checkmark & 55.7 & 10.1\\
\hline
\end{tabular}
}
\vspace{1pt}
\caption{Ablation study on the GTA5-to-Cityscapes adaptation. ST: the basic self-training method without any fusions. TF: temporal fusion by consistency loss. CMF: cross-model fusion by jointly generating offline pseudo labels. OOF: online-offline fusion through online pseudo label supervision.}
\vspace{-6pt}
\label{tab:result_method}
\end{wraptable}

Table~\ref{tab:result_method} investigates the contribution of each component of MFA on GTA5-to-Cityscapes. In the first row, ``Warm'' is the best warm-up model based on domain alignment methods~\cite{RN180,RN135}. ``ST'' is the basic self-training ( without any information fusion). We divide MFA to three key components, \emph{i.e.}, the temporal fusion (TF), the cross-model fusion (CMF) and the online-offline pseudo label fusion (OOF). Accordingly, we draw three observations. First, self-training brings $+3.8\%$ mIoU improvement over the warm-up training. Such improvement is consistent with many other two-stage UDA methods \cite{RN135,RN132,RN180}. Second, all the three components are important for MFA. By incrementally adding the key components, the performance reached 50.6\%, 51.7\% and 55.7\%, respectively. Third, these three fusion strategies adds up to +6.3\% mIoU improvement compared to basic ST. They jointly enable MFA achieve significant superiority against other two-stage UDA methods.

\section{Conclusion}
We propose a self-training method named Multi-Fusion Adaptation (MFA) for domain adaptive semantic segmentation. Through a co-learning framework, MFA integrates three information fusion strategies, \emph{i.e.}, cross-model fusion, temporal fusion, and online-offline pseudo label fusion. These fusions jointly suppress the pseudo label noises and explore informative samples during the self-training procedure. Consequentially, MFA significantly improves adaptive semantic segmentation and sets new state of the art on two popular benchmarks. 

\paragraph{Acknowledgments.}
This work was supported by the National Natural Science Foundation of China under Grant 61802380, the National Key Research and Development Program of China under Grant 2019YFB1405100 and the Strategic Priority Research Program of the Chinese Academy of Sciences under Grant No. XDA19020500.
\bibliography{egbib}
\end{document}